\documentclass[10pt,twocolumn,letterpaper]{article}

\usepackage{3dv}
\usepackage{times}
\usepackage{epsfig}
\usepackage{graphicx}
\usepackage{amsmath}
\usepackage{amssymb}
\usepackage{caption}
\usepackage{comment}
\usepackage[position=top]{subcaption}
\usepackage[export]{adjustbox} 
\usepackage{enumitem} 

\usepackage[normalem]{ulem}

\usepackage{xcolor}

\makeatletter
\newcommand{\citemissing}[1]{%
  \def\tmp{#1}%
  \ifx\tmp\empty
    \textcolor{red}{\textbf{[\#REF]}}%
  \else
    \textcolor{red}{\textbf{[\#REF: #1]}}%
  \fi%
}
\newcommand{\missing}[1]{%
  \def\tmp{#1}%
  \ifx\tmp\empty
    \textcolor{red}{\textbf{XYZ}}%
  \else
    \textcolor{red}{\textbf{#1}}%
  \fi%
}
\makeatother

\definecolor{mike_comment}{rgb}{0.1,0.5,0.1}

\definecolor{xavier_comment}{rgb}{0.5,0,0.8}

\newcommand\sbullet[1][.5]{\mathbin{\vcenter{\hbox{\scalebox{#1}{$\bullet$}}}}}
\newcommand\textlidarpoint[2]{$\!$\textcolor{#2}{$\sbullet[1.5]$}\hspace{-0.62em}\textcolor{#1}{$\sbullet[.9]$}}
\definecolor{ground_color}{rgb}{0.05,0.43,0.77}
\newcommand{\groundpoint}[0]{\textlidarpoint{ground_color}{black}}
\definecolor{nonground_color_1}{rgb}{0.31,0.89,0.61}
\definecolor{nonground_color_2}{rgb}{0.92,0.60,0.60}
\definecolor{nonground_color_3}{rgb}{0.98,0.80,0.61}
\newcommand{\nongroundpoints}[0]{%
  \textlidarpoint{nonground_color_1}{black}, %
  \textlidarpoint{nonground_color_2}{black}, %
  \textlidarpoint{nonground_color_3}{black}%
}
\newcommand{\primaryarrow}[0]{\protect{\includegraphics[scale=1]{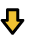}}}
\newcommand{\secondaryarrow}[0]{\protect{\includegraphics[scale=1]{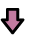}}}

\usepackage{bbding}
\usepackage{pifont}

\usepackage{mathtools}
\DeclarePairedDelimiter\floor{\lfloor}{\rfloor}

\newlength{\savedfboxsep}

\usepackage[numbers]{natbib} 
\usepackage{gensymb} 

\usepackage[pagebackref=true,breaklinks=true,letterpaper=true,colorlinks,bookmarks=false]{hyperref}

\threedvfinalcopy 

\def\threedvPaperID{188} 

\title{Simulated LiDAR Repositioning : a novel point cloud data augmentation method}

\author{{Xavier Morin-Duchesne}
\\
McGill University\\
Canada\\
{\tt\small xavier.morinduchesne@mail.mcgill.ca}
\and 
{Michael S. Langer}
\\
McGill University\\
Canada\\
{\tt\small langer@cim.mcgill.ca}
}

\begin{document}
\maketitle

\begin{abstract}
We address a data augmentation problem for LiDAR.  Given a LiDAR scan of a scene from some position, how can one simulate new scans of that scene from different, secondary positions? The method defines criteria for selecting valid secondary positions, and then estimates which points from the original point cloud would be acquired by a scanner from these positions. We validate the method using synthetic scenes, and examine how the similarity of generated point clouds depends on scanner distance, occlusion, and angular resolution. We show that the method is more accurate at short distances, and that having a high scanner resolution for the original point clouds has a strong impact on the similarity of generated point clouds. We also demonstrate how the method can be applied to natural scene statistics: in particular, we apply our method to reposition the scanner horizontally and vertically, separately consider points belonging to the ground and to non-ground objects, and describe the impact on the distributions of distances to these two classes of points.
\end{abstract}


\section{Introduction}

In this paper, we pose a simple problem: given a 3D LiDAR scan obtained from one position, how can one simulate new scans from different, secondary positions? We present a method, {\it Simulated LiDAR Repositioning} (SLR), which solves this problem and demonstrate its properties. We also show how it can be used in the context of natural scene statistics research.

At a high level, the method is straightforward. First, choose a set of candidate secondary positions from which to simulate scanning; these can be chosen manually or automatically based on some criteria. Then, for each of these secondary positions, define the viewing coordinates and the angular resolution. Finally, for each secondary position, solve the hidden surface problem~\cite{Sutherland}, and obtain a point cloud built from a subset of the samples of the original point cloud. This is a data augmentation method: it generates new point clouds representing different views of existing point clouds.

We envision several use cases for our data augmentation method. The first is to generate large numbers of point clouds from existing ones by repositioning the scanner horizontally along the ground surface. This can be useful since LiDAR scanner are expensive, acquiring LiDAR data is time consuming, and many of the publicly available datasets feature a limited number of point clouds---\ie between 1 and 100~\cite{SYNS, Semantic3D, PandaSet, Paris-Lille, Paris-rue-Madame, Toronto3D, iQmulus, TUMMLS}. The second use case is to change the scanner's height relative to the ground. This could be useful, for example, in autonomous vehicle research. LiDAR scanners that are mounted on vehicles are always at the same height above the ground and this may introduce biases in the sampling of the ground surface (see Section~\ref{sec:distancedistributions}). Varying scanner height can reduce this bias. A third use case that we envisage is with Mobile Laser Scanning (MLS) system datasets. MLS point cloud datasets typically consist of individual scan point clouds, aggregate point clouds comprising all of the scans of a given scene, or both. Our method allows one to use the aggregate point clouds in order to simulate placing a scanner somewhere in the scene. This is useful for studies employing static (non-Mobile)  scanner datasets, since this essentially transforms MLS point clouds into static scanner point clouds.

Our SLR method was originally devised in the context of {\it natural scene statistics} (NSS) research. The idea behind NSS is that the brain adapts to the statistics of the environment, whether through efficient coding of image properties ~\cite{simoncelli2001natural}, or through a process of Bayesian inference where the priors are statistics of the environment~\cite{geisler2008visual}. While most studies to date have focused on RGB images, others have used point clouds to study NSS \cite{howe2002range, huang2000rangeimages, YangPurves2003} with some even combining the two by using co-registered RGB and point clouds or range images~\cite{SYNS, cooper2014perceived, PotetzLee03,  su2017bayesian}. It was in fact while revisiting work by Yang and Purves~\cite{YangPurves2003} that we realized how useful having a large number of lower resolution scans from many different viewpoints might be. Yang and Purves extracted distance NSS from point clouds to explain certain well-known biases in visual distance perception. We were interested in reproducing some of their findings on newer datasets. We present some of our findings using point clouds from the SYNS~\cite{SYNS} dataset in Section~\ref{sec:distancedistributions} as an example application of our method.

Point clouds and co-registered RGB and range images have also frequently been used by the computer vision community. The NYU indoor scene database~\cite{NYU} and the KITTI autonomous vehicle database~\cite{KITTI}, for example, are commonly used to train deep learning algorithms. The KITTI dataset in particular has often been used to train networks to estimate depth from RGB images with the low resolution LiDAR providing the true distance. This highlights another potential use case for our method: generating more low resolution LiDAR scans for training depth estimation models. This also highlights another important point: while this paper focuses on applications of SLR to NSS research and related applications, we believe that the our method may be applicable to areas of computer vision and machine learning which employ 3D point clouds~\cite{Guo:20}, and more generally has the potential to be useful in many fields which use 3D point clouds.

The rest of the paper is organized as follows. In Section~\ref{sec:SLR}, we describe the steps of our Simulated LiDAR Repositioning (SLR) method in detail. In Section~\ref{sec:SLR_Validation}, we present an experiment to validate the SLR method using synthetic scenes, and show that similarity between true scans and simulated scans depends on the distance from the original scanner, occlusion (\ie the number of objects in the scene), and the angular resolution of the original scan. In Section~\ref{sec:distancedistributions}, we apply our method to NSS research problems. In particular, we expand on some of Yang and Purves'~\cite{YangPurves2003} analyses using SLR and distinguishing between points belonging to the ground (ground points) and those belonging to non-ground objects (non-ground points). We find that distances to ground and to non-ground points are distributed differently, and that those distributions are impacted differently by viewing height. Finally, we conclude in Section~\ref{sec:conclusion}.

\section{Simulated LiDAR Repositioning Method}
\label{sec:SLR}

The SLR method has three main steps. First, one or more secondary scan positions within the point cloud are selected; we employed a grid-based method to identify potential secondary scan positions, and selected from among those. Second, an angular resolution is chosen for the simulated LiDAR scanner; it must be lower than that of original scanner and appropriate given the point density at the new position. Third, for each selected secondary position, an SLR point cloud is created by subsampling the point cloud to appear as if the scanner had been positioned at the  secondary position.

We  describe these steps in greater detail below. These assume that points have been labelled as ground or non-ground. Some point cloud datasets already come with such labels; semantic labeling \cite{Semantic3D} and autonomous vehicle research \cite{PandaSet,iQmulus,TUMMLS}, for example, typically offer some form of point labelling which includes ground point labelling. For unlabelled datasets, one can use a method such as the Cloth Simulation Filtering~\cite{CSF} to label ground versus non-ground points, and indeed that is what we do in the experiments in Section \ref{sec:distancedistributions}.

\subsection{Selecting a secondary position}
\label{ssec:method:grid_selection}

We propose a grid-based approach for automatically identifying potential secondary (simulation) positions.  Given a point cloud, the idea is to cover it with a grid of cells, eliminate cells which don't meet the criteria described below, and to select secondary positions---\ie positions at which to simulate a scan---from within the remaining cells. This approach is optional; others may wish to use a different approach for selecting secondary positions, or may wish to select them manually. Supplementary Figure~\ref{supp:fig:grid_selection} provides a helpful representation of the result of each step, applied to a point cloud from the iQumulus \& TerraMobilita MLS dataset~\cite{iQmulus}.

First, the $xy$ (horizontal) plane is covered with a grid of $2\times{}2$~m cells, and cells that do not have any points in them are removed (Figure~\ref{supp:fig:grid_selection:b}).

Then, cells with an insufficient number of ground points in their vicinity are also removed (Figure~\ref{supp:fig:grid_selection:c}). The average vertical position of ground points in the vicinity of the scanner is used during the subsampling step (Section~\ref{ssec:method:SLR:subsampling}) to determine the vertical position of the simulated scanner. In order for this estimate of ground height to be reliable, a large number of these points is needed. Our experiments led us to reject any cells which had less than 50\,000 ground points in a 5~m radius from its center, and to use the average vertical position of these points within a 3~m radius during subsampling. While these parameters worked well for our datasets, they could be modified depending on scene sampling density.

Next, cells with an insufficient number of points in the various azimuth directions are removed. The intuition behind this step is twofold. First, we want to remove cells partially occluded by large objects for which many azimuth directions would have few or even no points (\eg a cell around a corner). Second, at the same time, we want to  allow some azimuth directions to have a low number of points as may occur in natural scene datasets. Since there does not exist a formula to determine the number of points that an azimuth direction should have, the following \textit{ad hoc} method is used. First, a {\it minimum profile} is defined based on the scans from a reference, natural scene dataset. For each scan in this dataset, points are divided between 360 one-degree-azimuth bins regardless of their elevation, and the bin counts are sorted from low to high. Then, the minimum profile is defined by comparing the sorted histograms across scans, and taking the minimum value of each bin. (Since each histogram is sorted, so is the minimum profile.) Next, a similar profile (a sorted one-degree-azimuth histogram) is created for each cell, and compared to the minimum profile. Cells for which any of the sorted bin counts are less than those of the minimum profile are removed (Figure~\ref{supp:fig:grid_selection:d}). In this paper, we use Yang and Purves'~\cite{YangPurves2003} dataset as the reference dataset.

Finally, secondary positions are chosen by randomly selecting cells from those remaining, and randomly choosing a position within each of those cells. Note that the criteria defined above were selected based on datasets with which we were working, and may need to be adjusted for use with other datasets.

\subsection{Choosing the angular resolution}

Once a secondary position has been selected, one must choose an appropriate angular resolution for the simulated scanner. As one moves away from the original position of the scanner, the density of points drops. Consequently, to simulate LiDAR repositioning in a position away from the original position, an angular resolution appropriate for the point density at the new position must be chosen. We consider two ways of doing this: one can calculate an angular resolution appropriate for the distance away from the original scanner position, or one can use the angular resolution of the scanner from the reference dataset used to create the minimum profile described in Section~\ref{ssec:method:grid_selection}. Note that the second method is more easily applicable and can be applied to MLS dataset point clouds. (The first is more complicated to apply, if it can be at all, since there is no single original scanner in an MLS point cloud.)
For our purposes, we used azimuth and zenith angular resolutions of 0.144\degree{}, the same as that of the scanner from our reference dataset~\cite{YangPurves2003}.

\subsection{Simulating a LiDAR scan}
\label{ssec:method:SLR:subsampling}

The crux of the SLR method is to subsample the original point cloud in order to simulate what an ideal static LiDAR scanner does, in particular, deciding which points are considered to be visible from the secondary scanner position.
To measure distances, LiDAR scanners fire laser pulses and capture their reflections. The time of flight of each pulse is used to determine the distance of the object off of which the pulse bounced. 
The volume covered by an ideal LiDAR scanner can be divided into pyramids where each pyramid represents the 3D space within which a single  pulse would detect an object. For simplicity, we assume that, should there be an object in this space, the pulse will bounce off the closest portion of that object and return. For our simulations, this means that, if there is a single point within a pulse's range, that point is returned, representing the single object detected by the pulse. Should there be multiple points within a pulse's range, corresponding to multiple returns from one or more objects, only the closest point is returned (\ie the closest portion of the closest object); other points within the pulse's range are discarded.  Computing which is the closest 3D point in each pulse's range basically amounts to solving the hidden surface removal problem of computer graphics \cite{Sutherland}.

The process of picking points and mimicking LiDAR scanning is as follows. Point cloud coordinates are first transformed from Euclidean to spherical coordinates; we used the~$(R, \theta, \phi)$ form where $R$ = radius, $\theta$ = zenith angle, and $\phi$ = azimuth. Then, for each point, spherical coordinates are transformed to binned coordinates using

\begin{equation}
    (R, v, h) = (R, \floor*{\frac{\theta}{\theta_{res}}}, \floor*{\frac{\phi}{\phi_{res}}})
\end{equation}
where $\theta_{res}$ and $\phi_{res}$ are the simulated vertical and horizontal angular resolutions, and $\floor{\ }$ is the floor operator. Points are then sorted by $h$, $v$, and $R$, in that order. Finally, all that remains is to iterate over the sorted list of binned coordinates, picking the first point in each bin and discarding the others. This has the advantage of only considering combinations of $v$ and $h$ for which there are samples and is rather efficient: point clouds in the SYNS outdoor dataset have around 29\,000\,000 points on average, and subsampling those point clouds takes an average of 18.64s (and no more than 32s) on a single 2.3GHz core. The result is a new point cloud simulating what a LiDAR scanner might acquire from the new position.

\subsection{Limitations}
\label{sssec:method:lidar_simulation_limitations}

Before turning to the experiments, we discuss two limitations of the SLR method. First, the method can only represent points that were present in the original point cloud, and does not attempt to infer back-facing points or occluded points which were not visible during the original scan. Consider a forest scene, for example.  The method cannot simulate points from the back side of a tree since these points were not captured by the original LiDAR scanner. The tree would still be present in the SLR point cloud, but it would be represented by points from its front side irrespective of the position from which the new point cloud was created.
Note that our method does attempt to address this occlusion problem
indirectly using the criteria outlined in Section~\ref{ssec:method:grid_selection}.
For example, one of the criteria is that there should be
a large number of ground points in the vicinity of potential secondary positions, and another is that there should be
a certain number of points at each azimuth direction.
Both of these aim to remove secondary positions around corners or behind large objects.

The second limitation has to do with two biases inherent to any single view imaging method like LiDAR scanning or capturing RGB images: sampling bias and positioning bias. The former is a bias where points near the scanner are sampled more densely, and the latter is one where the data may be biased because the scans were taken from an arbitrary point chosen by the user. As a result, objects that were sampled more densely in the original point cloud are more likely to be visible in the point clouds generated by the method, and objects that were not sampled, of course, will simply not appear. Note that, while our method does not avoid these biases entirely, it does mitigate these biases: the biases introduced by one scan from a scene are diminished by having a large number of simulated scans from the same scene.

\section{Experiment 1: SLR Validation}

\label{sec:SLR_Validation}

Our first experiment addresses three questions about the method itself. First, how is the similarity between SLR point clouds and to their true counterparts affected by distance from the original scanner? Second, how does occlusion impact the similarity between SLR point clouds and their true counterpart? Third, how does the original point cloud's angular resolution affect similarity? We validate our method by answering these three questions and showing that the method remains accurate at further distances and despite occlusion.

\subsection{Method}

\begin{figure*}[t]
  \begin{center}
    \includegraphics[scale=0.9]{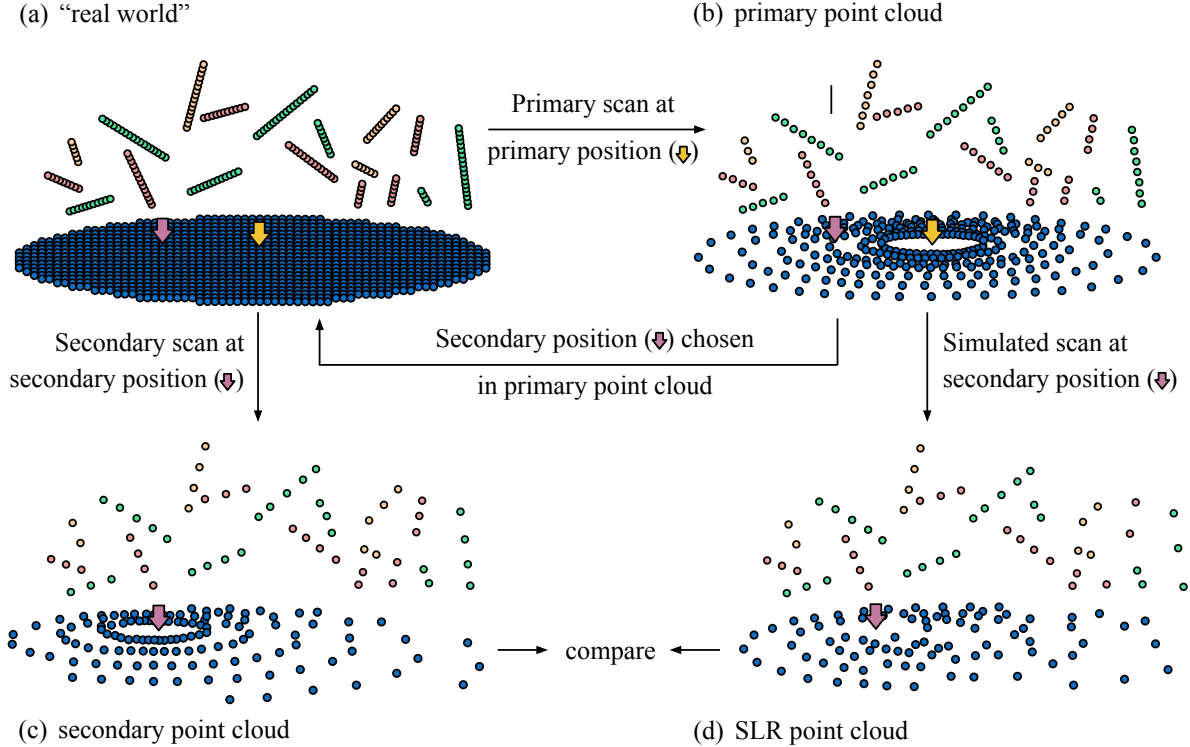}
    \phantomsubcaption
    \label{fig:validation:dense}
    \phantomsubcaption
    \label{fig:validation:primary}
    \phantomsubcaption
    \label{fig:validation:secondary}
    \phantomsubcaption
    \label{fig:validation:simulated}
  \end{center}
  \caption{
    Validating Simulated LiDAR Repositioning (SLR). Dark blue points (\groundpoint{}) are used to represent points belonging to the ground plane, whereas green, red, and orange points (\nongroundpoints{}) are used to represent point belonging to non-ground objects. Non-ground objects are represented in 1D for simplicity.
    \textbf{(a)} A dense point cloud representing a ``real world'' scene is created.
    \textbf{(b)} A real-world scan is simulated from the center of the dense point cloud (\primaryarrow{}) by applying the simulation step of our SLR method from the center of the dense point cloud (``real world''). The resulting point cloud is intended to be similar to that from LiDAR point cloud datasets. This point cloud has a high angular resolution (0.0144\degree{} or 0.00576\degree{}).
    \textbf{(c)} Secondary positions are selected using the primary point cloud based on criteria described in text. A secondary point cloud is created by simulating acquiring a point cloud from secondary positions in the ``real world,'' that is to say by applying the subsampling step of our SLR method at each secondary position in the dense point cloud. Secondary point clouds have a low angular resolution (0.144\degree{}).
    \textbf{(d)} SLR point clouds are created by applying the simulation step of our SLR method at each secondary position in the primary point cloud. SLR point clouds have the same, low angular resolution as secondary point clouds (0.144\degree{}). Furthermore, whereas secondary point clouds represent what one \textit{would} acquire at the secondary position (true scan), SLR point clouds represent what one \textit{might} acquire (simulated scan). Secondary and SLR point clouds are compared to address three questions about our method.
  } 
  \label{fig:validation}
\end{figure*}

Figure~\ref{fig:validation} illustrates our validation method. The steps are as follows. First, we create a dense point cloud which represents the real, continuous world (Figure~\ref{fig:validation:dense}). Second, we simulate a scan in this point cloud and obtain a primary point cloud (Figure~\ref{fig:validation:primary}) representing a scan from a static LiDAR scanner in the real world. Third, we use the criteria outlined in Section~\ref{ssec:method:grid_selection} to identify potential secondary positions. Fourth, we simulate a scan at a subset of these positions in the dense point cloud, and obtain secondary point clouds (Figure~\ref{fig:validation:secondary}); these represent what a scanner in a secondary positions \textit{would} acquire (true scan). Fifth, we also simulate a scan at the same subset of positions in the primary point cloud, and obtain SLR point clouds (Figure~\ref{fig:validation:simulated}); these represent what a scanner in a secondary positions \textit{might} acquire (simulated scan). Finally, we compare each secondary point cloud to the SLR point cloud which was created from the same secondary position. Each step is described in greater detail below. The last step, the comparison, is discussed in the next section.

For the first step, creating a dense point cloud representative of the real world (Figure~\ref{fig:validation:dense}), we generate a round disk with a radius of 100~m and composed of points along a Euclidean grid with a spacing of 5~mm (40\,000 points/m${}^2$). This disk represents the ground surface. We position it at $z = -1.65$. (Static LiDAR scanners typically have a height of 1.65~m.) We also generate between 0 and 2000 planar rectangles with the same point density---see details below. These planar rectangles are meant to represent non-ground surfaces which a LiDAR scanner might also detect. The planar rectangles' size, position, and orientation are randomly chosen. Size varies between 0.1~m and 10.1~m, $x$ and $y$ coordinates between -100~m and 100~m, and $z$ between -5~m and 10~m. Orientation is uniformly randomly sampled. Points below $z = -1.65$~m and outside of the 100~m radius are removed.

To answer the three questions described at the beginning of Section~\ref{sec:SLR_Validation}, we repeat the first step a total of nine times, creating nine dense point clouds. For the first question, five dense point clouds are generated with 1000 planar rectangles each; an average of 754 planar rectangles are present or partially present after removing points below the ground surface or outside the 100~m radius. As for the second and third questions, four dense point clouds are generated with 0 (ground surface only), 500, 1000, and 2000 planar rectangles, resulting in 0, 388, 749, and 1500 at least partially present planar rectangles.

For the second step, simulating a scan in the real world (Figure~\ref{fig:validation:primary}), we use the scan simulation step of the SLR method (Section~\ref{ssec:method:SLR:subsampling}) with the dense point cloud, positioning the simulated scanner in the center of the dense point cloud at a height of $z = 0$. For the first question (about distance and similarity) and the second questions (occlusion and similarity), the simulated scanner has an angular resolution of 0.0144\degree{}. For the third question (angular resolution and similarity), a finer angular resolution of 0.00576\degree{} is used.

\begin{figure}
    \centering
    \includegraphics[scale=0.33]{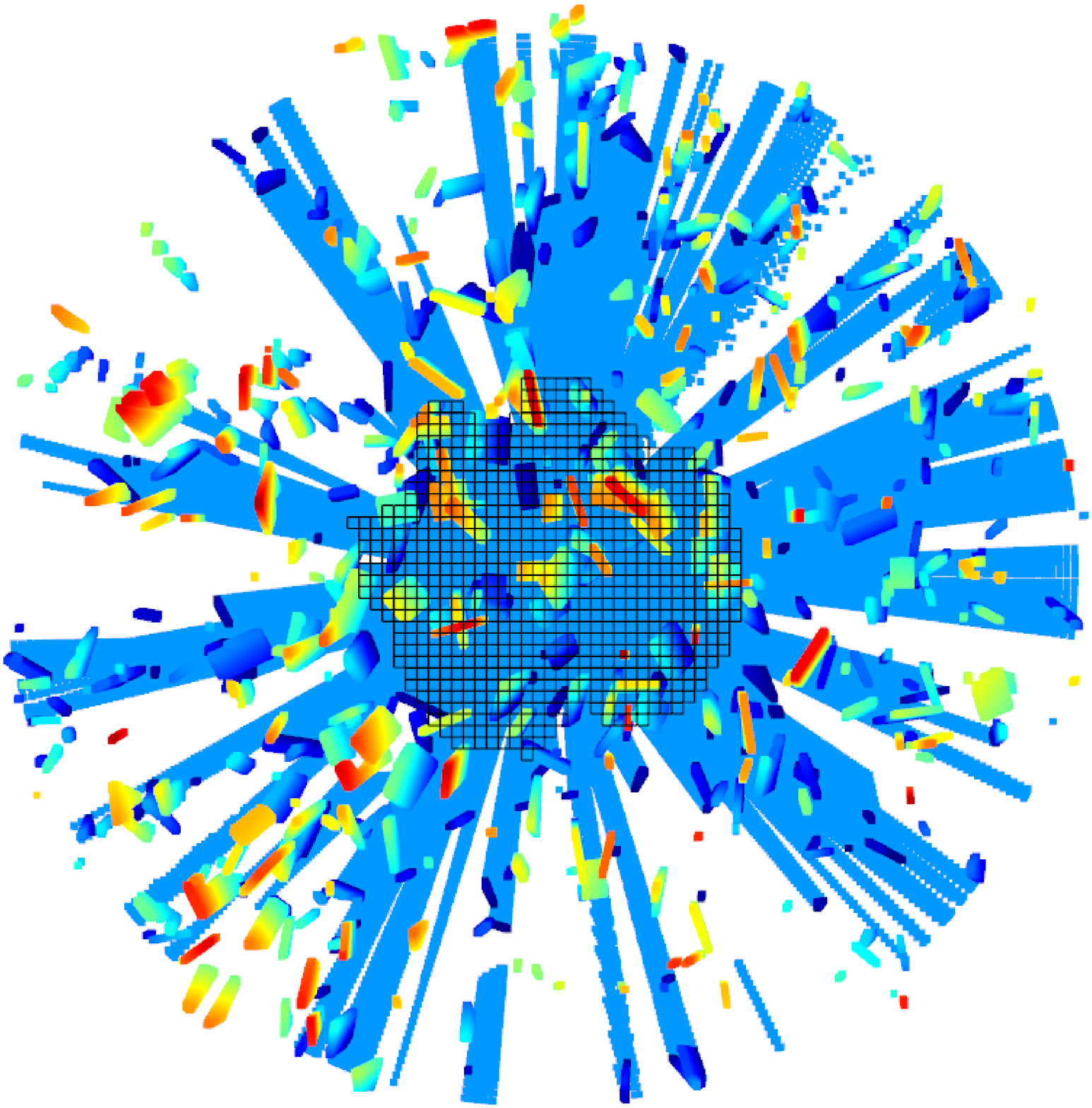}
    \caption{Example primary point cloud viewed from above. The grid shows the cells from which secondary positions were chosen. This primary point cloud was created from a dense point cloud with 1000 planar rectangles, and has an angular resolution of 0.0144\degree{}}
    \label{fig:primary_cloud_grid}
\end{figure}

For the third step, we apply the grid-based approach and criteria from the SLR method (Section~\ref{ssec:method:grid_selection}) to the primary point clouds, obtain a large number of potential secondary positions, and randomly select a subset of 100 among those.
More specifically, we obtain an average of 724 potential secondary positions for the five point clouds for the first question, and 963, 763, 730, and 574 secondary positions for the 0, 500, 1000, and 2000 planar rectangle point clouds for the second question. From each, we select 100 secondary positions, resulting in 900 secondary positions. For the third question, we use the same subset of secondary positions as for the second question. Figure~\ref{fig:primary_cloud_grid} shows a primary point cloud and its grid of cells from which secondary locations were chosen.


For the fourth and fifth steps, simulating real-world scans and simulating SLR scans at secondary positions (Figures~\ref{fig:validation:secondary} and~\ref{fig:validation:simulated}), we apply the scan simulation step of the SLR method (Section~\ref{ssec:method:SLR:subsampling}) at each selected secondary position to the dense point cloud (fourth step) and the corresponding primary point cloud (fifth step) respectively. The simulated scanners have an angular resolution of 0.144\degree{} (10 and 25 times coarser than those of the second step) for both secondary and SLR point clouds, and for all three questions. This results in 100 secondary-SLR pairs for each primary point cloud, where each secondary point cloud represents what a scanner \textit{would} acquire at a given position (true scan), and its corresponding SLR point cloud represents what a scanner \textit{might} acquire at the same position (simulated scan).

The code was written in Python~\cite{Python3} using Numpy~\cite{Numpy}, Pandas~\cite{pandas}, and Open3D~\cite{Open3D}. Dense point clouds were quite large, with an average of 948\,000\,000 points. While the SLR method does not have particularly high memory requirements, due to the size of the dense point clouds, the code for this experiment was run on Compute Canada clusters with large amounts of memory.

\begin{figure*}[t]
    \centering
    \leftskip-0.2cm
    \begin{subfigure}[t]{0.02\textwidth}
        \subcaption{}
        \label{fig:exp1:1000}
    \end{subfigure}
    \begin{subfigure}[t]{0.305\textwidth}
        \includegraphics[valign=t, scale=1]{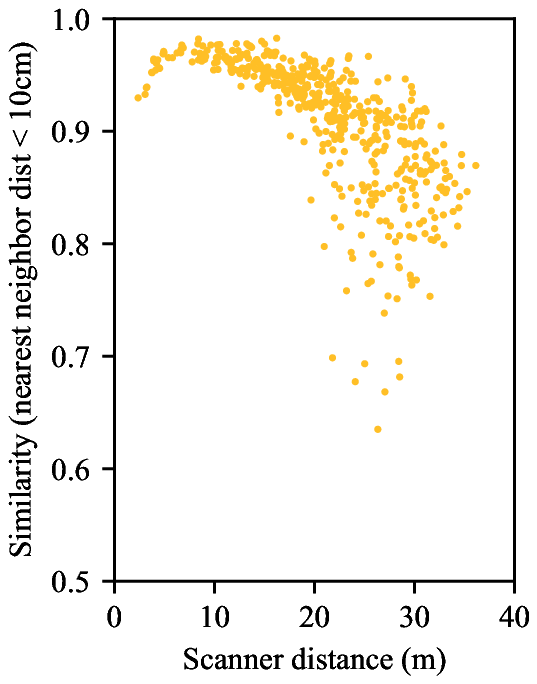}
    \end{subfigure}
    \begin{subfigure}[t]{0.02\textwidth}
        \subcaption{}
        \label{fig:exp1:occlusion}
    \end{subfigure}
    \begin{subfigure}[t]{0.305\textwidth}
        \includegraphics[valign=t, scale=1]{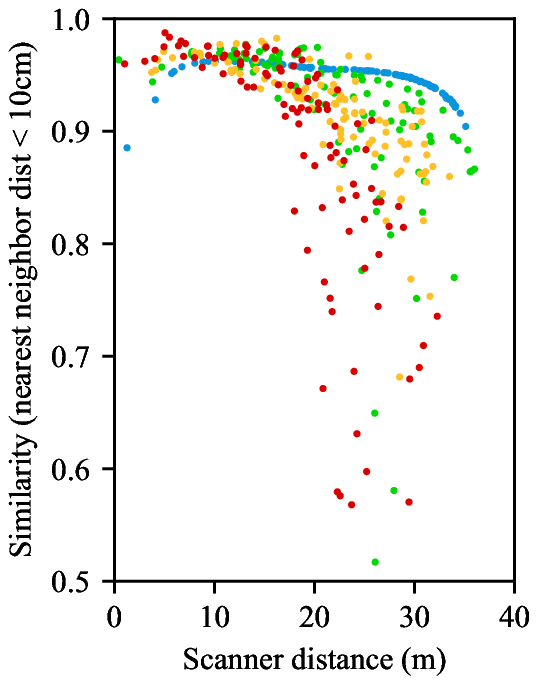}
    \end{subfigure}
    \begin{subfigure}[t]{0.02\textwidth}
        \subcaption{}
        \label{fig:exp1:angular_resolution}
    \end{subfigure}
    \begin{subfigure}[t]{0.305\textwidth}
        \includegraphics[valign=t, scale=1]{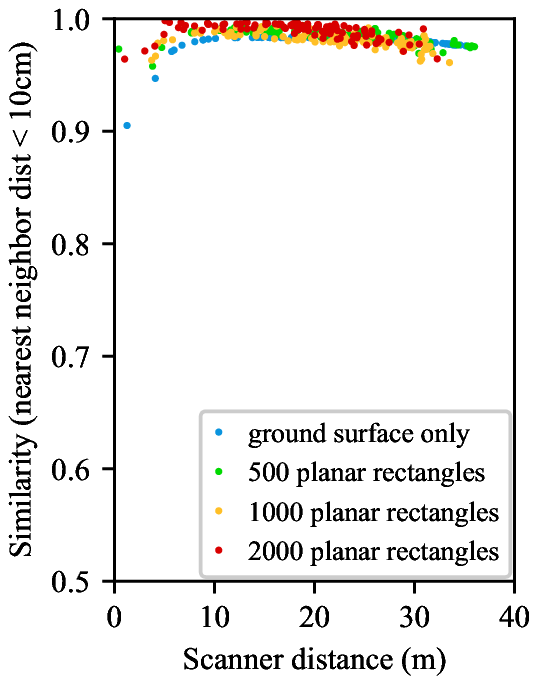}
    \end{subfigure}
    \caption{
        Results from Experiment 1 on method validation.
        \textbf{(a)} Scatterplot of the proportion of points in a secondary point cloud that have a nearest neighbor distance of less than 10 cm to their SLR point cloud, as a function of the distance between the secondary and primary positions.
        \textbf{(b)} Same plot as in (a), but with varying numbers of planar rectangles (non-ground objects) in each ``real world'' scene (dense point cloud) which affects the level of occlusion.
        \textbf{(c)} Same comparison as in (b), but the primary point clouds have finer angular resolution.
    }
    \label{fig:results}
\end{figure*}

\subsection{Results \& Discussion}

We set out to answer three questions: How is similarity between SLR point clouds and their true counterparts affected by distance from the original scanner? How does occlusion impact the similarity between SLR point clouds and their true counterparts? How does the original point cloud's angular resolution affect similarity?

To answer the first, regarding similarity and distance, we use the 500 secondary-SLR point cloud pairs created from the five 1000 planar rectangle dense point clouds and the five corresponding primary point clouds with an angular resolution of 0.0144\degree{}. In particular, given a secondary-SLR point cloud pair, we define {\it similarity} between the two point clouds to be the proportion of points from the secondary point cloud (true scan) which have their \textit{nearest neighbor} in the SLR point cloud less than 10~cm away. We compute nearest neighbor distance using an Open3D library~\cite{Open3D} function called {\small\texttt{compute\_point\_cloud\_distance}}. We also compute \textit{scanner distance}---\ie the distance from the secondary position (\secondaryarrow{}) to the primary position (\primaryarrow{}; Figure~\ref{fig:validation})---for each pair. Similarity is plotted as a function of scanner distance in Figure~\ref{fig:exp1:1000}. As expected, the method is most accurate at short scanner distances: for scanner distances below 17.5~m, over 90\% of points in the secondary point cloud had their closest neighbor in the SLR point cloud less than 10~cm away. As for scans beyond 17.5~m, the method is often still very accurate. In more than 92\% of secondary-SLR point cloud pairs, over 80\% of the secondary point cloud's points have a nearest neighbor distance less than 10~cm to the SLR point cloud.

For the second question, about the impact of occlusion, we use the 400 secondary-SLR point cloud pairs created from the dense point clouds with 0, 500, 1000, and 2000 planar rectangles and the primary point clouds with an angular resolution of 0.0144\degree{}. Similarity and scanner distance are computed in the same way. Occlusion is defined as the number of planar rectangles in the dense point cloud. The results are presented in Figure~\ref{fig:exp1:occlusion}. As expected, at  large scanner distances,  similarity diminishes as occlusion  increases (\ie more objects in the scene). Interestingly, occlusions appear however to have little-to-no impact at short scanner distances. 

As for the third question, about the impact of the original (primary) point cloud's resolution, we use the same 400 secondary point clouds from the previous question, but use the corresponding SLR point clouds created from the primary point cloud with a finer angular resolution of 0.00576\degree{}. Similarity and scanner distance are again computed in the same way, and the result is shown in Figure~\ref{fig:exp1:angular_resolution}. Interestingly, compared to Figures~\ref{fig:exp1:1000} and~\ref{fig:exp1:occlusion}, there appears to be little-to-no impact of scanner distance. The most likely explanation is that, by making the original point cloud's angular resolution 2.5 times finer, we extended the range of scanner distances at which the method is accurate.

In conclusion, the method's accuracy appears to depend on scanner distance, on occlusion, and, most of all, on the angular resolution of the original point cloud. In particular, given a high enough angular resolution for the original point cloud, the method appears to be very accurate even at large distances away from the original scanner position.



\section{Experiment 2: Distance distributions }
\label{sec:distancedistributions}

The second set of experiments applies the SLR method to a  problem of natural scene statistics. We revisit a study by Yang and Purves \cite{YangPurves2003} that examined distance distributions of LiDAR scanned scenes. One of their basic findings was that distance histograms tended to peak at around 3~m. They argued that this finding was consistent with a bias in human vision to perceive surfaces to be at around this distance, in the sense of a Bayesian prior that is based on natural scene statistics. We investigate whether this finding applies equally to distances to points belonging to the ground (ground points), and to points belonging to non-ground objects (non-ground points). We also reexamine Yang and Purves' analysis on the impact of viewing height, use our SLR method to simulate scanners at different heights, and consider the impact of this manipulation on distances to the ground and to non-ground objects.

\subsection{Method}

We present examples for one dataset only:  76 outdoor scenes from the  Southampton-York Natural Scenes (SYNS) dataset \cite{SYNS}. (The results are very similar for other data sets tested, see~\cite{in_preparation}). The point clouds were captured from both fully-natural scenes (\eg forests) and scenes with natural and man-made elements (\eg mall, beach with cottages). One scan was captured per scene, and the SLR method was used to augment these data.

The scanner height for the SYNS dataset was always 1.65~m above the ground. For scan directions, the vertical range was from the zenith to 135\degree{} away from the zenith, leaving a 45\degree{} elevation hole where the scanner base was. The maximum distance range of the scanner was 120 m and the angular resolution was 0.036\degree{} in both zenith (elevation) and azimuth.

We labeled points from the original scans as ground or non-ground using the cloth Simulation Filtering (CSF) method~\cite{CSF}. This was achieved by modifying the code for the CSF plugin provided with CloudCompare~\cite{CloudCompare}. The experiment was otherwise carried out using code written in Python~\cite{Python3}, and employed the Numpy~\cite{Numpy}, Pandas~\cite{pandas}, and Open3D~\cite{Open3D} libraries. The code was executed on Compute Canada clusters, and on a desktop computer with a 12-core 2.3GHz CPU and 32GB of RAM. One secondary position was selected for each cell, and one point cloud was generated for each position for a total of 13\,000 SLR point clouds across all of the SYNS outdoor point clouds. While our implementation of the subsampling step is fairly efficient (less 20s per secondary position), given the large number of generated point clouds, executing the analyses still took several days of computational time.

\captionsetup[subfigure]{singlelinecheck=off}
\begin{figure}[b]
    \leftskip-1em
    \begin{subfigure}[t]{0.25\textwidth}
        \raggedleft
        \subcaption{\leftskip1cm Original point clouds}
        \label{fig:exp2:raw_vs_sim:raw}
        \includegraphics[scale=1,valign=t]{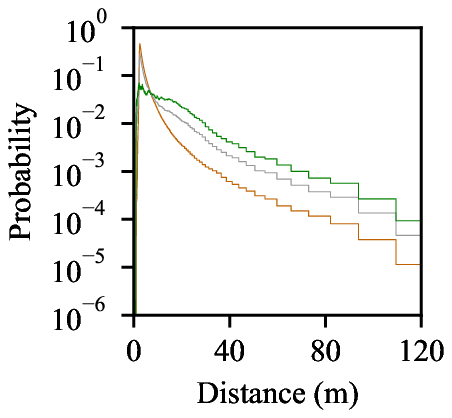}
    \end{subfigure}
    \begin{subfigure}[t]{0.20\textwidth}
        \subcaption{\leftskip0.55cm SLR point clouds}
        \label{fig:exp2:raw_vs_sim:sim}
        \includegraphics[scale=1,valign=t]{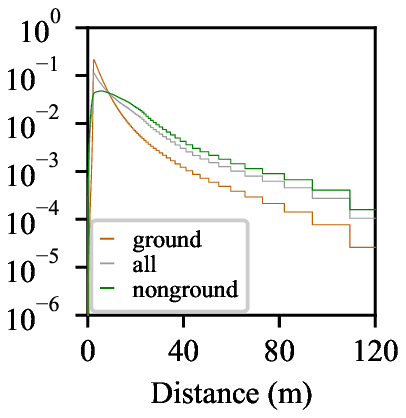}
    \end{subfigure}\\[1em]
    \begin{subfigure}[t]{0.25\textwidth}
        \raggedleft
        \subcaption{\leftskip1cm Ground}
        \label{fig:exp2:height:ground}
        \includegraphics[scale=1,valign=t]{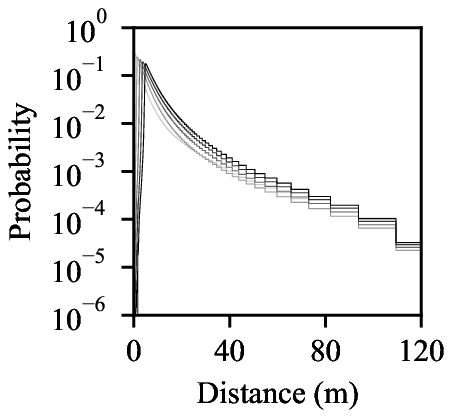}
    \end{subfigure}
    \begin{subfigure}[t]{0.20\textwidth}
        \subcaption{\leftskip0.55cm Non-ground}
        \label{fig:exp2:height:nonground}
        \includegraphics[scale=1,valign=t]{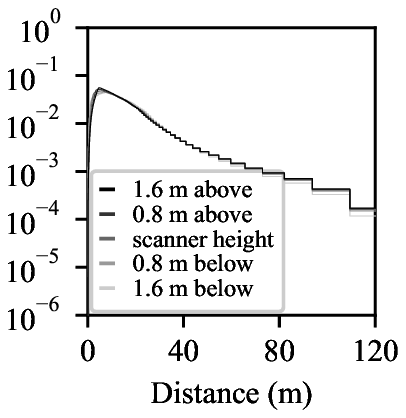}
    \end{subfigure}
    \caption{
        Results from Experiment 2 on applying our SLR method to distance NSS research problems. Details in text.
        \textbf{(a)} Probability distributions for distances to all points, ground points, and non-ground points in the SYNS outdoor dataset~\cite{SYNS}.
        \textbf{(b)} The same probability distributions in SLR point clouds created from the SYNS dataset.
        \textbf{(c)} Probability distributions for distances to ground points at different scanner heights.
        \textbf{(d)} Probability distributions for distances to non-ground points at different scanner heights.
        Note that the observed effects---particularly those in \subref{fig:exp2:raw_vs_sim:raw} and \subref{fig:exp2:raw_vs_sim:sim}---cannot simply be explained by the proportion of ground points: ground points represented 39.16\% of all points on average ($SD = $ 16.54\%) in the SYNS outdoor scenes.
    }
    \label{fig:exp2}
\end{figure}

In the next section, distance histograms are plotted using \textit{pulse-sized bins,} namely distance bins the width of which corresponds to the distance range covered by the pulse of an ideal LiDAR scanner sitting at 1.65~m above a pure ground plane. Note that the bin sizes increase with distance both because of the pulse size increases with distance (an angle effect) and because of increased foreshortening of the ground as distance increases.

\subsection{Results}

Figure~\ref{fig:exp2} presents distributions of all distances, distances to the ground, and distances to non-ground objects. 
Figure~\ref{fig:exp2:raw_vs_sim:raw} shows the pooled data over the original 76 point clouds, and Figure~\ref{fig:exp2:raw_vs_sim:sim} shows the data for the point clouds created using SLR. The distributions of distances to all points concord with Yang and Purves' findings: they peak around 3~m, and decline in a similar fashion. In contrast, the distributions of distances to the ground versus non-ground objects peak at different locations and have different shapes. This difference is more evident in the SLR point cloud data plot, which is smoother because it pools over more scans. This suggests that distributions of distances to ground points and to non-ground points may not equally concord with the psychophysical evidence for distance perception biases.

We were also interested in Yang and Purves' analysis on the impact of viewing height on horizontal distances. In particular, they investigated the way in which the height of an observer might influence the way in which distances are distributed in natural environments (\eg a child vs a giant's perspective). Here, we extend Yang and Purves' analysis in two ways. We use our SLR method to manipulate scanner height, and separately consider distances to the ground and non-ground objects.
Figures~\ref{fig:exp2:height:ground} and~\ref{fig:exp2:height:nonground} show our results for ground and non-ground points respectively.
The peaks of the ground distributions approach distance zero as the scanner is lowered towards the ground, which is what one might expect since the ground is closer in each pulse direction as the scanner height decreases (at least for ground points below the horizon). The distributions of distances to non-ground points are approximately identical across heights. These effects are more apparent in the data when the curves are sufficiently smooth. This requires that enough scans are available, which is where the SLR method pays off.

\section{Conclusion}
\label{sec:conclusion}

In this paper, we presented a data augmentation method for LiDAR: Simulated LiDAR Repositioning. This method allows one to take a (primary) LiDAR scan of a scene as input, and produces as output lower resolution (simulated) LiDAR scans from a set of secondary scan positions. We proposed several criteria for selecting secondary scan positions, such as requiring secondary positions to have a sufficient number of ground points in their vicinity or requiring that they have at minimum a number of points over the various azimuth directions.

We described many use cases for our data augmentation method: generating a large number of scans for a given scene from different horizontal positions, changing the vertical position of the scanner, and transforming MLS aggregate point clouds into static scanner point clouds. It is likely that there are other potential uses for it, especially in fields like machine learning where having a large quantity of data is important.

We presented two experiments. The first experiment validated the method using synthetic scenes. We showed how the similarity of the secondary and SLR point clouds depends on the distance between the secondary position and original (primary) scanner position, as well as on the number of objects in the scenes (because of occlusion effects). We also showed that the angular resolution of the primary point cloud is important: with a sufficiently high resolution, similarity between the secondary and simulated point clouds remains high even at larger distances.

The second experiment applied our method to natural scene statistic research. In particular, we extended earlier analyses by Yang and Purves \cite{YangPurves2003} using our method and a more recent dataset~\cite{SYNS}. We found that using a large number of generated point clouds provided similar but smoother distributions. We also discovered notable differences in the distributions of distances to the ground and to non-ground objects, especially when the height of the scanner is changed.   The  distribution of distances to ground points shifted with scanner height, whereas the distribution to non-ground points stayed constant. These results are encouraging, and suggest that our method could also be used to explore other natural scene statistics such a surface orientation \cite{SYNS} and more generally scene layout.

{\small
\bibliographystyle{ieee}
\bibliography{egbib}
}

\setcounter{table}{0}
\renewcommand{\thetable}{S\arabic{table}}
\setcounter{figure}{0}
\renewcommand{\thefigure}{S\arabic{figure}}

\begin{figure}
    \phantomcaption
    \label{supp:fig:grid_selection}
    \phantomsubcaption
    \label{supp:fig:grid_selection:b}
    \phantomsubcaption
    \label{supp:fig:grid_selection:c}
    \phantomsubcaption
    \label{supp:fig:grid_selection:d}
\end{figure}

\end{document}


\section*{Supplemental material}

\setcounter{table}{0}
\renewcommand{\thetable}{S\arabic{table}}
\setcounter{figure}{0}
\renewcommand{\thefigure}{S\arabic{figure}}
  
\begin{figure}[h!]
    \centering
    \begin{subfigure}[t]{0.03\textwidth}
        \subcaption{}
        \label{supp:fig:grid_selection:b}
    \end{subfigure}
    \begin{subfigure}[t]{0.74\textwidth}
        \fbox{\includegraphics[width=\linewidth, valign=t]{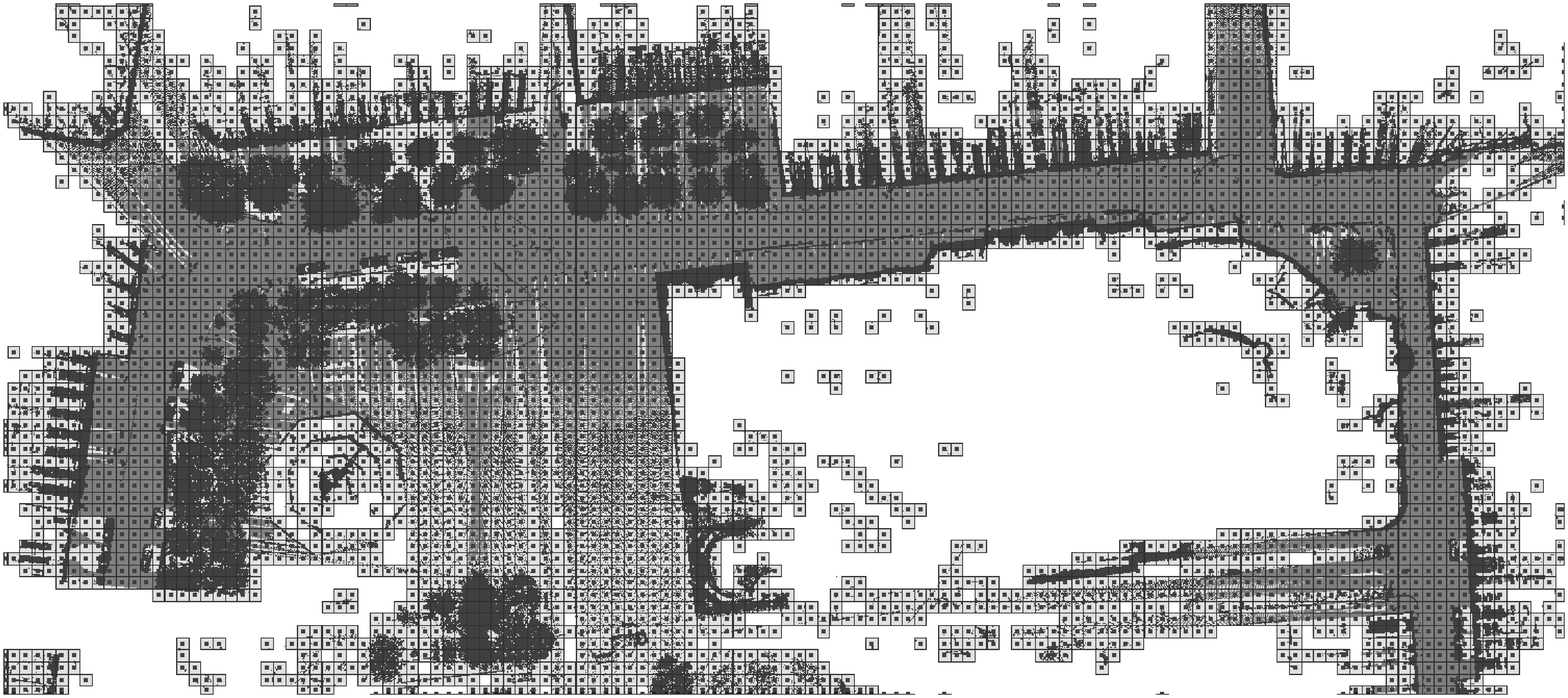}}
    \end{subfigure}\\
    \begin{subfigure}[t]{0.03\textwidth}
        \subcaption{}
        \label{supp:fig:grid_selection:c}
    \end{subfigure}
    \begin{subfigure}[t]{0.74\textwidth}
        \fbox{\includegraphics[width=\linewidth, valign=t]{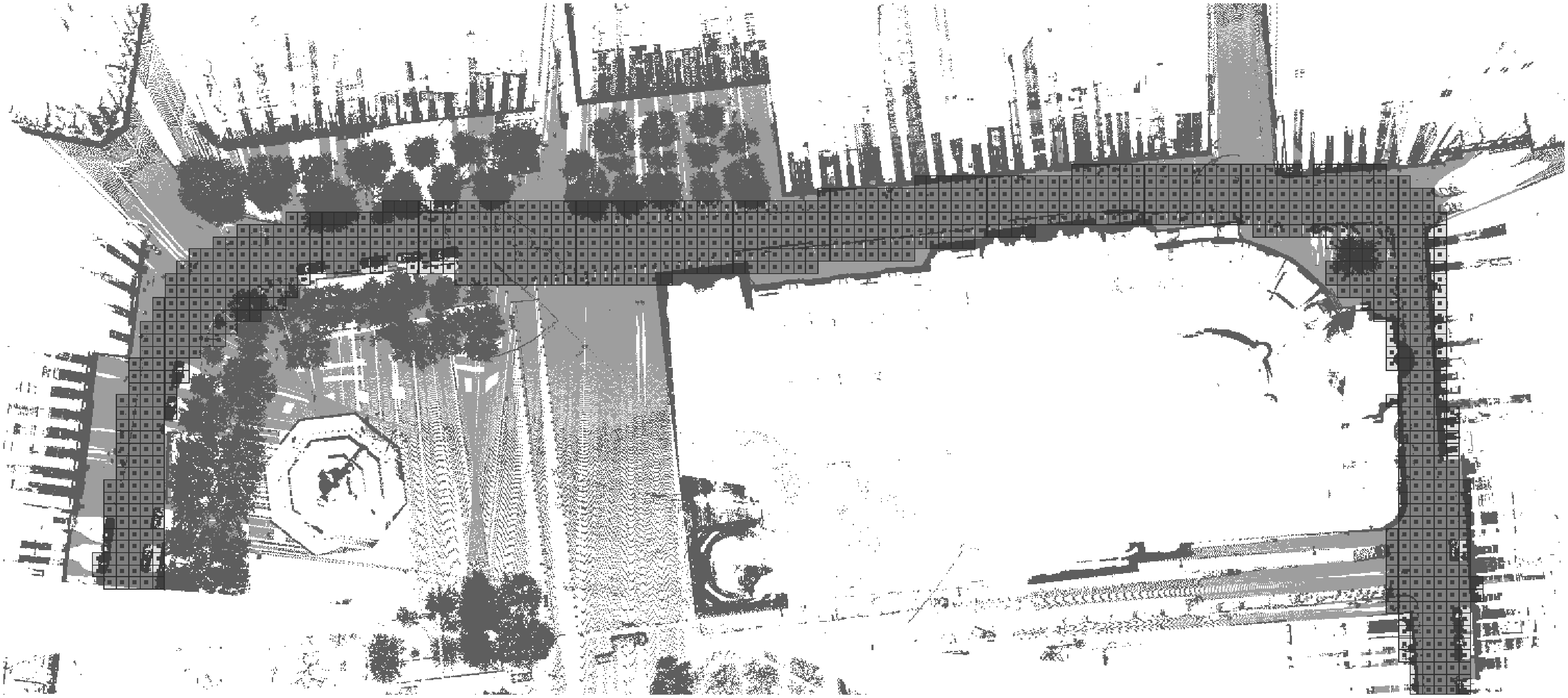}}
    \end{subfigure}\\
    \begin{subfigure}[t]{0.03\textwidth}
        \subcaption{}
        \label{supp:fig:grid_selection:d}
    \end{subfigure}
    \begin{subfigure}[t]{0.74\textwidth}
        \fbox{\includegraphics[width=\linewidth, valign=t]{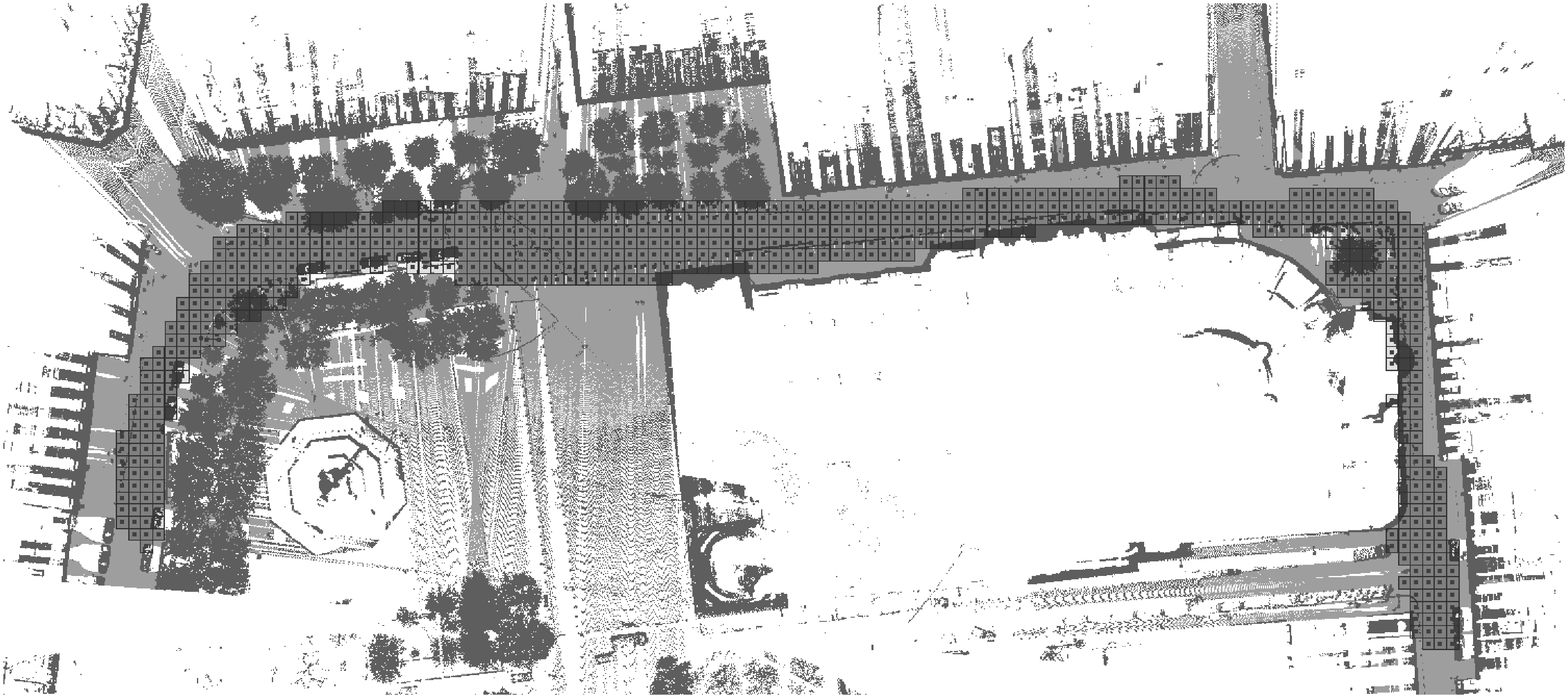}}
    \end{subfigure}
    \caption{
        View from the top of scene Z8 from the iQmulus \& TerraMobilita dataset showing the steps involved in selecting new positions for Simulated LiDAR Repositioning using a grid-based approach. In these figures, cells are identified by \protect\includegraphics[scale=0.75]{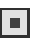}. Areas which do not present this symbol were removed.
        \textbf{(a)} A grid is created to cover the whole scene (not shown), and empty cells are then removed.
        \textbf{(b)} Next, for each cell, the number of ground points within a 5~m radius from the center of the cell is counted. If this number is below 50\,000, the cell is removed.
        \textbf{(c)} For each remaining cell, the number of points per degree along the azimuth is counted and sorted. The sorted counts must be at least as great as that of the minimum profile created from a reference dataset (Yang and Purves' dataset here).
    }
    \label{supp:fig:grid_selection}
\end{figure}